\documentclass[Afour,sageh,times]{sagej}

\usepackage{moreverb,url}

\usepackage[colorlinks,bookmarksopen,bookmarksnumbered,citecolor=blue,urlcolor=blue]{hyperref}
\usepackage[switch]{lineno} 
\usepackage{algorithm} 
\usepackage{algpseudocode}
\usepackage{makecell}
\usepackage{graphicx} 
\usepackage{multirow}

\newcommand\BibTeX{{\rmfamily B\kern-.05em \textsc{i\kern-.025em b}\kern-.08em
T\kern-.1667em\lower.7ex\hbox{E}\kern-.125emX}}

\def\volumeyear{2026}

\begin{document}

\runninghead{Jian et al.}

\title{Disentangling Damage from Operational Variability: A Label-Free Self-Supervised Representation Learning Framework for Output-Only Structural Damage Identification}

\author{Xudong Jian\affilnum{1}, Charikleia Stoura\affilnum{2}, Simon Scandella\affilnum{1}, Eleni Chatzi\affilnum{1}}

\affiliation{\affilnum{1}Department of Civil, Environmental and Geomatic Engineering, ETH Z\"urich, Z\"urich, Switzerland}

\affiliation{\affilnum{2}Department of Mechanical Engineering, Politecnico di Milano, Italy}

\corrauth{Xudong Jian, Department of Civil, Environmental and Geomatic Engineering, ETH Z\"urich, Z\"urich, Switzerland.}

\email{xudong.jian@ibk.baug.ethz.ch}

\begin{abstract}
Damage identification is a core task in structural health monitoring. In practice, however, its reliability is often compromised by confounding non-damage effects, such as variations in excitation and environmental conditions, which can induce changes comparable to or larger than those caused by structural damage. To address this challenge, this study proposes a self-supervised label-free disentangled representation learning framework for robust vibration-based structural damage identification.
The proposed framework employs an autoencoder with two latent representations to learn directly from raw vibration acceleration signals. A self-supervised invariance regularization, implemented via Variance–Invariance–Covariance Regularization (VICReg), is imposed on one latent representation using baseline data where structural damage is assumed constant but operational and environmental conditions vary. In addition, a frequency-domain constraint is introduced to enforce agreement between the power spectral density reconstructed from the latent representation and that computed from the corresponding input time series. Together, these mechanisms promote disentanglement, enabling the learned representation to be sensitive to damage-related characteristics while remaining invariant to nuisance variability.
The framework is trained in a fully end-to-end and label-free manner, requiring no prior information on damage, excitation, or environmental conditions, making it well-suited for real-world applications. Its effectiveness is validated on two distinct real-world vibration datasets, including a bridge and a gearbox. The results demonstrate robustness to operational variability, strong generalization capability, and good performance in both damage detection and quantification.
\end{abstract}

\keywords{Structural health monitoring, structural damage identification, deep learning, self-supervised learning, disentangled representation learning, operational variability}

\maketitle

\section{Introduction}
Engineering structures such as bridges are essential to societal and economic activities. As these assets age and face increasing demands in terms of loads, ensuring their safety and serviceability is of paramount importance \cite{an2019recent}. Structural health monitoring (SHM) has therefore attracted sustained attention over the past decades as a means to support structural condition assessment using in-situ measurements and data interpretation \cite{farrar2007introduction}.

A typical SHM workflow comprises data acquisition, signal processing, and decision-making for downstream tasks, including damage detection, localization, quantification, and prognosis. From a data acquisition perspective, in the civil infrastructure context, vibration-based SHM is among the most widely-adopted paradigms \cite{carden2004vibration}. This is attributed to the strong correlation between the dynamic response of a structure and its structural parameters, such as stiffness, mass, and boundary conditions \cite{doebling1996damage}. From a signal processing perspective, SHM methods are commonly grouped into physics-based approaches, data-driven approaches, and hybrid approaches that combine the former two. The vast use of SHM in practice has increased data volume and complexity, making data-driven and hybrid approaches, particularly those based on deep learning, of greater interest \cite{farrar2012structural}.

However, a major obstacle for reliable SHM using data-driven and hybrid approaches is the presence of confounding non-damage influences. In practical scenarios, variations in operational and environmental conditions (for example, traffic intensity, wind, temperature, humidity, and changing excitation characteristics) can alter structural properties as much as, or more than, actual damage \cite{sohn2007effects,li2021time}. This phenomenon is widely recognized as a central challenge in SHM and is frequently discussed in terms of operational and environmental variability, or, in machine learning terms, as a distribution shift across monitoring conditions \cite{figueiredo2011machine}. In principle, additional measurements and explicit compensation models can mitigate these effects \cite{peeters2001one}, but comprehensive measurement of all relevant non-damage factors is rarely feasible for real-world structures \cite{reynders2014output}. In practice, a robust SHM method should therefore be able to separate damage-sensitive information from nuisance variability even when excitation and environmental factors are only partially observed or not observed at all.

To develop such robust SHM methods, a broad range of research has been conducted. Early research in this regard relies on feature engineering and statistical normalization of damage-sensitive indicators. Representative efforts include output-only strategies that seek stable features under changing conditions, and time series-based approaches that explicitly remove non-damage trends from monitored features \cite{deraemaeker2008vibration,reynders2014output}. In particular, cointegration-based methods aim to construct combinations of measured responses or extracted features that remain approximately stationary under benign variability, so that deviations are more likely attributable to damage \cite{cross2011cointegration}. Regime-switching extensions further attempt to handle multiple operating regimes by modelling piecewise-stationary behaviour \cite{shi2018regime}. While effective in some settings, these approaches typically rely on linearity or weak nonlinearity assumptions, require careful selection of features and sensors, and may struggle when nuisance effects are high-dimensional, non-stationary, or strongly coupled with damage.

Those limitations caused by manual feature selection and assumptions motivate the use of deep learning to directly extract useful representations from raw vibration signals. In this regard, unsupervised deep learning has been widely explored for damage detection, most commonly through autoencoders and related reconstruction-based anomaly detection schemes \cite{eltouny2023unsupervised}. Extensions include integrating information fusion and active sensing concepts to enhance anomaly detectability \cite{wang2026unsupervised}, or combining autoencoders with filtering frameworks to explicitly neutralize temperature effects \cite{cadini2022neutralization}. However, purely reconstruction-driven representations can still entangle damage with nuisance variability, and models trained under one set of operating conditions may exhibit degraded reliability when monitoring conditions shift, which motivates the need for learning strategies that explicitly target robustness to distribution shifts.

To address distribution shifts more directly, domain adaptation and domain-aware learning have been introduced into SHM to align feature distributions across environments and operating regimes, enabling more transferable damage classifiers. Examples include domain adaptation applied to bridge monitoring for damage classification across varying conditions \cite{giglioni2024domain}, as well as approaches that fuse damage-sensitive features with domain adaptation to improve robustness on real building datasets \cite{martakis2023fusing}. More recent work also considers open-set settings where unknown damage classes may appear under varying environments, using open-set domain adaptation to improve reliability under both environmental variability and class uncertainty \cite{zhou2024structural}. Despite these advances, many domain adaptation pipelines still depend on labelled source damage data and may not explicitly separate damage-related factors from nuisance factors in a disentangled manner, which becomes a bottleneck when labels are scarce and when quantification or interpretation of the learned representation is desired.

Therefore, recent studies have begun to incorporate self-supervised learning and disentanglement concepts into SHM feature learning. For example, contrastive and self-supervised learning have been used to enhance open-set damage recognition when only healthy data and limited minor damage data are available \cite{duthe2024towards,zhou2025transferring,shi2025contrastive}. Domain separation has also been explored through architectures such as domain-separated capsule networks \cite{jiang2025domain}, and physics-informed variational autoencoders have been proposed to disentangle confounding sources by augmenting latent spaces with interpretable physics-based variables and regularizing the learning objective \cite{koune2024disentangled}. In parallel, self-supervised strategies have been explored at the population level to enhance damage detection and model tuning across similar structures \cite{bel2025structural}. These efforts collectively support the importance of self-supervision and disentanglement, but they may still require domain labels, partial supervision, or explicit physics models, and they do not typically impose both baseline invariance and accessible physics on a dedicated damage-sensitive latent representation, leaving room for a simpler end-to-end framework that relies only on measured vibration time series.

To address the above-mentioned limitations, this study proposes a label-free, self-supervised, disentangled representation learning framework for robust vibration-based structural damage identification under non-stationarity, nonlinearity, and varying operational conditions. The framework is built on an autoencoder with two latent representations, designed to disentangle damage-sensitive information from nuisance variability using only measured vibration time series, without requiring any damage, excitation, or environmental labels during training. The main contributions of this study are as follows:

\begin{enumerate}
    \item We introduce a self-supervised invariance regularization on one latent representation, denoted $\mathbf{z}_\mathrm{dmg}$, using Variance Invariance Covariance Regularization (VICReg) \cite{bardes2022vicreg}. The regularization is applied on signal windows drawn from a baseline period where the structural damage state is assumed constant while operational and environmental conditions vary. This encourages $\mathbf{z}_\mathrm{dmg}$ to remain consistent within the baseline state without requiring excitation or environmental measurements.
    
    \item We propose a frequency-domain loss that constrains $\mathbf{z}_\mathrm{dmg}$ across all conditions by matching the power spectral density (PSD) decoded from $\mathbf{z}_\mathrm{dmg}$ with the PSD computed from the corresponding vibration signal windows. This physics-guided constraint improves spectral fidelity and promotes retention of damage-relevant frequency characteristics without using any labels.
    
    \item We systematically evaluate the proposed framework on two real vibration datasets, including a full-scale bridge experiment and a high-fidelity gearbox experiment. The evaluation includes multiple train validation test splits, ablation studies, and latent space analysis using UMAP \cite{mcinnes2018umap}, demonstrating robustness to operational variability and improved damage detection and quantification performance.
\end{enumerate}

\section{Methodology}
\subsection{Problem statement}
This study focuses on structural damage detection and quantification using only vibration acceleration measurements. This setting is representative of many practical SHM applications, where direct measurements of excitation and environmental conditions are unavailable.

As discussed earlier, measured vibration signals contain not only information related to structural damage, but also contributions from non-damage factors, such as operational and environmental variations. The objective of this study is therefore to disentangle these two types of information in a learned latent space. To this end, a deep learning model is employed to map raw vibration signals to two latent representations:
\begin{equation}\label{eq:problem1}
\left( \mathbf{z}_{\mathrm{dmg}},\, \mathbf{z}_{\mathrm{ndmg}} \right) = f_\theta \left( \mathbf{X} \right),
\end{equation}
where $\mathbf{X} \in \mathbb{R}^{C \times T}$ denotes a signal window extracted from the measured acceleration time series, with $C$ sensing channels and $T$ time samples. The function $f_\theta(\cdot)$ represents a neural network parameterized by $\theta$, which learns latent representations directly from the input data. The vectors $\mathbf{z}_{\mathrm{dmg}} \in \mathbb{R}^{H}$ and $\mathbf{z}_{\mathrm{ndmg}} \in \mathbb{R}^{H}$ denote $H$-dimensional ($H \ll T$) latent representations corresponding to damage-sensitive and non-damage-related information, respectively.

Based on the learned damage-sensitive representation $\mathbf{z}_{\mathrm{dmg}}$, a scalar damage score is defined as
\begin{equation}\label{eq:problem2}
m = g \left( \mathbf{z}_{\mathrm{dmg}} \right),
\end{equation}
where $g(\cdot)$ denotes a mapping from the latent space to a scalar damage indicator $m \in \mathbb{R}$. This metric can then be used for both damage detection (e.g., via thresholding) and damage quantification (e.g., via relative magnitude comparison).

\subsection{Model architecture}
\begin{figure*}[!htp] 
\centering
\includegraphics[width=12.1cm]{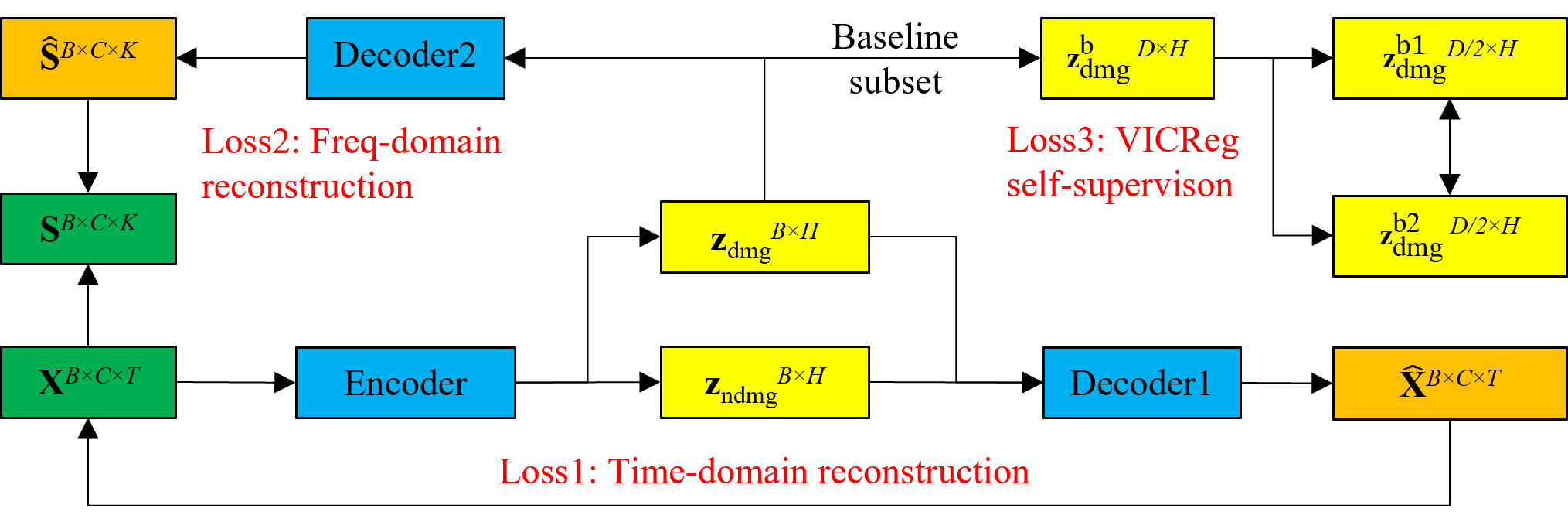}
\caption{Proposed self-supervised disentangled representation learning architecture. A batch of input vibration signals $\mathbf{X}$ (batch size: $B$) is encoded into a damage-sensitive latent representation $\mathbf{z}_{\mathrm{dmg}}$ and a nuisance latent representation $\mathbf{z}_{\mathrm{ndmg}}$. The model is trained with three objectives: (1) time-domain reconstruction of $\mathbf{X}$, (2) frequency-domain reconstruction to match the PSD, and (3) self-supervised invariance regularization (VICReg) on $\mathbf{z}_{\mathrm{dmg}}$ using baseline pairs. }
\label{fig:arch}
\end{figure*}

Building on prior studies \cite{eltouny2023unsupervised,cadini2022neutralization,koune2024disentangled}, this study designs the deep learning model architecture as illustrated in Figure \ref{fig:arch}.

As shown in Figure \ref{fig:arch}, the proposed architecture consists of three main modules:

\begin{itemize}
    \item \textbf{Encoder}: The encoder processes the input acceleration time series $\mathbf{X}$ and maps them into two latent representations, $\mathbf{z}_{\mathrm{dmg}}$ and $\mathbf{z}_{\mathrm{ndmg}}$, corresponding to damage-sensitive and nuisance-related information, respectively. This design enables the disentanglement of structural damage effects from operational and environmental variability in the latent space. Since the focus of this study is not on network architecture innovation, a one-dimensional convolutional neural network (1D CNN), which is simple yet effective for time series modelling, is adopted as the encoder.

    \item \textbf{Decoder 1}: Decoder 1 reconstructs the input time series $\mathbf{\hat{X}}$ from the concatenated latent representations $\left[\mathbf{z}_{\mathrm{dmg}}, \mathbf{z}_{\mathrm{ndmg}}\right]$. The reconstruction objective is to ensure that the latent space retains sufficient information from the original signals, enabling the model to learn compact and informative representations. A 1D CNN is again used for Decoder 1, which is standard for time-domain signal reconstruction.

    \item \textbf{Decoder 2}: Decoder 2 takes only the damage-sensitive latent representation $\mathbf{z}_{\mathrm{dmg}}$ as input and reconstructs the normalized power spectral density $\mathbf{\hat{S}}$ corresponding to the input signal $\mathbf{X}$. This frequency-domain reconstruction introduces a frequency-domain constraint that encourages $\mathbf{z}_{\mathrm{dmg}}$ to preserve damage-relevant spectral characteristics. Since this task does not require temporal convolution, a multi-layer perceptron (MLP) is employed for Decoder 2.
\end{itemize}

For brevity, the detailed mechanisms and architectural configurations of the 1D CNN and MLP are not presented herein, as they constitute standard algorithms in the deep learning field. Nonetheless, all implementation details, including network structures and hyperparameters, are provided in our GitHub repository, which has been made publicly available to ensure the reproducibility of this study.

\subsection{Training strategies}
As discussed earlier, a key practical challenge in applying deep learning to structural damage identification is the scarcity of labeled data. In most real-world SHM scenarios, labels related to structural damage, excitation, and environmental conditions are either very limited or entirely unavailable, which makes supervised training difficult and may degrade model performance. To address this challenge, the proposed framework adopts three complementary loss functions (training objectives), as illustrated in Figure \ref{fig:arch}. Their details are described below:

\begin{itemize}
    \item \textbf{Loss 1: time-domain reconstruction loss}. This loss function, which is fundamental for training autoencoders, minimizes the discrepancy between the reconstructed acceleration time series $\mathbf{\hat{X}}$ and the measured signals $\mathbf{X}$. By enforcing accurate reconstruction, the model is encouraged to automatically learn informative representations from raw data. The loss is defined as:
\begin{equation}\label{eq:loss1}
\mathcal{L}_1=\frac{1}{CT}\sum_{c=1}^{C}\sum_{t=1}^{T}\left(x_{c,t}-{\hat{x}}_{c,t}\right)^2
\end{equation}

\item \textbf{Loss 2: frequency-domain reconstruction loss}. This loss function minimizes the difference between the reconstructed power spectral density $\mathbf{\hat{S}}$ derived from $\mathbf{z}_{\mathrm{dmg}}$ and the magnitude-normalized power spectral density $\mathbf{S}$ computed from the input signals $\mathbf{X}$. The motivation is to encourage $\mathbf{z}_{\mathrm{dmg}}$ to capture frequency-domain characteristics that are closely related to structural properties and comparatively less sensitive to excitation variability. The loss is defined as:
\begin{equation}\label{eq:loss2}
\mathcal{L}_2=\frac{1}{CK}\sum_{c=1}^{C}\sum_{k=1}^{K}\left(s_{c,k}-{\hat{s}}_{c,k}\right)^2
\end{equation}
where $K$ is the number of data points of power spectral density.
    
    \item \textbf{Loss 3: self-supervised VICReg loss}. 
This loss function is designed to regularize the damage-sensitive latent representation $\mathbf{z}_{\mathrm{dmg}}$ in a self-supervised manner, with the objective of suppressing sensitivity to operational and environmental variability while preserving damage-related information.

Specifically, as shown in Figure \ref{fig:arch}, for each training batch of latent representations $\mathbf{z}_{\mathrm{dmg}} \in \mathbb{R}^{B \times H}$, a subset of $D$ samples ($D < B$), denoted as $\mathbf{z}^b_{\mathrm{dmg}} \in \mathbb{R}^{D \times H}$, is selected from a baseline period, during which the structural state is assumed to remain unchanged, while excitation and environmental conditions may vary. From these $D$ samples, two groups are constructed by selecting latent vectors with odd and even indices, respectively, forming paired representations $\left(\mathbf{z}^{b1}_{\mathrm{dmg}}, \mathbf{z}^{b2}_{\mathrm{dmg}}\right) \in \mathbb{R}^{(D/2) \times H}$. These pairs are treated as \textit{positive pairs} that share the same structural state but differ in nuisance factors.

To enforce invariance while avoiding representation collapse, Variance–Invariance–Covariance Regularization (VICReg) \cite{bardes2022vicreg} is adopted. The overall loss is defined as:
\begin{equation}\label{eq:loss3}
\mathcal{L}_3 = \lambda_{\mathrm{inv}} \, \mathcal{L}_{\mathrm{inv}} 
+ \lambda_{\mathrm{var}} \, \mathcal{L}_{\mathrm{var}} 
+ \lambda_{\mathrm{cov}} \, \mathcal{L}_{\mathrm{cov}}
\end{equation}
where the three terms are defined as follows:

\textbf{(1) Invariance term:}
\begin{equation}
\mathcal{L}_{\mathrm{inv}} = \frac{1}{H} \sum_{h=1}^{H} 
\frac{1}{D/2} \sum_{d=1}^{D/2} 
\left( z^{b1}_{d,h} - z^{b2}_{d,h} \right)^2
\end{equation}
This term minimizes the distance between paired representations, encouraging $\mathbf{z}_{\mathrm{dmg}}$ to be invariant to nuisance variations within the baseline condition.

\textbf{(2) Variance term:}
\begin{equation}
\begin{split}
\mathcal{L}_{\mathrm{var}} &=
\frac{1}{H} \sum_{h=1}^{H} \max\left(0, 1 - \sigma_h(z^{b1}) \right) \\
 &+
\frac{1}{H} \sum_{h=1}^{H} \max\left(0, 1 - \sigma_h(z^{b2}) \right)
\end{split}
\end{equation}
where $\sigma_h(\cdot)$ denotes the standard deviation of the $h$-th latent dimension across the batch. This term prevents collapse of latent dimensions by ensuring sufficient variability in each feature.

\textbf{(3) Covariance term:}
\begin{equation}
\mathcal{L}_{\mathrm{cov}} = 
\frac{1}{H} \sum_{i \neq j} 
\left( \mathrm{Cov}(z^{b1}_{i,j})^2 + \mathrm{Cov}(z^{b2}_{i,j})^2 \right)
\end{equation}
where $\mathrm{Cov}(\cdot)$ denotes the covariance matrix computed across the batch. This term penalizes correlations between different latent dimensions, encouraging feature decorrelation and improving representation expressiveness.

In this study, the weighting coefficients are set to $\lambda_{\mathrm{inv}} = 25$, $\lambda_{\mathrm{var}} = 25$, and $\lambda_{\mathrm{cov}} = 1$, following the original VICReg setting \cite{bardes2022vicreg} that provides the best performance in extensive experiments.
\end{itemize}

The overall training objective is defined as a weighted combination of the three loss terms:
\begin{equation}\label{eq:loss_all}
\mathcal{L} = \lambda_1 \, \mathcal{L}_1 
+ \lambda_2 \, \mathcal{L}_2 
+ \lambda_3 \, \mathcal{L}_3
\end{equation}

It is worth noting that $\mathcal{L}_1$ and $\mathcal{L}_2$ correspond to fully unsupervised objectives, requiring no assumptions or prior knowledge of structural damage, excitation, or environmental conditions. In contrast, $\mathcal{L}_3$ introduces a self-supervised constraint based on a baseline period, during which the structural state is assumed to remain unchanged while operational and environmental conditions can vary. This assumption is practical for most in-service structures, where damage typically evolves gradually rather than occurring abruptly.

\subsection{Damage identification}
After effective training, the encoder output $\mathbf{z}_{\mathrm{dmg}}$ is expected to primarily capture damage-sensitive information while being insensitive to nuisance variability. This representation is therefore used for structural damage identification. In this study, the Mahalanobis distance computed from $\mathbf{z}_{\mathrm{dmg}}$ is adopted as the damage score for both detection and quantification.

Let $\mathbf{z}_{\mathrm{dmg}} \in \mathbb{R}^{N \times H}$ denote the latent representations obtained from $N$ signal windows. Similar to the baseline subset used in the self-supervised learning introduced in Section "Training strategies" above, a subset corresponding to a baseline period is selected using a mask, denoted as $\mathbf{z}^b_{\mathrm{dmg}} \in \mathbb{R}^{N_b \times H}$, where $N_b$ is the number of baseline samples. It should be noted that the structure does not need to be strictly undamaged during the baseline period. Instead, a weaker and more practical assumption is adopted: the structural damage state remains unchanged during the baseline period, while excitation and environmental conditions may vary.

The mean vector $\boldsymbol{\mu}$ and covariance matrix $\boldsymbol{\Sigma}$ of the baseline distribution are computed by:
\begin{equation}
\boldsymbol{\mu} = \frac{1}{N_b} \sum_{n=1}^{N_b} \mathbf{z}^b_n, \quad
\boldsymbol{\Sigma} = \mathrm{Cov}(\mathbf{z}^b).
\end{equation}

The Mahalanobis distance for each sample $z_n$ is then computed by:
\begin{equation}
m_n = (\mathbf{z}_n - \boldsymbol{\mu})^\top \boldsymbol{\Sigma}^{-1} (\mathbf{z}_n - \boldsymbol{\mu}),
\end{equation}
yielding a scalar damage score for each sample.

The Mahalanobis distance provides a statistically grounded measure of deviation from the baseline distribution while accounting for correlations among latent features. Compared with Euclidean distance, it normalizes each dimension according to its variance and removes redundancy due to correlated features. This is particularly important in the learned latent space, where feature scales and dependencies are generally non-uniform, making the Mahalanobis distance well suited for this problem \cite{deraemaeker2018comparison,duthe2024towards}.

Damage detection is then formulated as a binary classification problem based on the damage scores. A threshold is determined from the baseline data as the $p$-th percentile (with $p=95$ in this study) of the baseline Mahalanobis distances:
\begin{equation}
\tau = \mathrm{Percentile}_p \left( \{ m_n \mid n \in \text{baseline} \} \right).
\end{equation}
Samples with $m_n > \tau$ are classified as damaged, while those with $m_n \leq \tau$ are classified as undamaged.

This percentile-based thresholding provides a fully data-driven and label-free decision rule, which is particularly suitable for SHM scenarios with limited or no damage labels. The resulting predictions can be evaluated using standard metrics such as accuracy and true positive rate. Moreover, the magnitude of $m_n$ serves as a continuous damage indicator, enabling damage quantification and progression analysis.

Finally, a pseudo-code is provided in Algorithm \ref{alg:damage_id} to clearly illustrate the damage identification procedure.

\begin{algorithm}[t]
\caption{Damage identification using Mahalanobis distance in latent space}
\label{alg:damage_id}
\begin{algorithmic}[1]

\Require Vibration signal windows $\{\mathbf{X}_n\}_{n=1}^{N}$; trained encoder $f_\theta(\cdot)$; baseline mask $\mathbf{b} \in \{0,1\}^N$; percentile $p$
\Ensure Damage scores $\{m_n\}_{n=1}^{N}$; threshold $\tau$; predictions $\{\hat{y}_n\}_{n=1}^{N}$

\vspace{0.3em}
\State \textbf{Step 1: Encode signals into latent space}

\State $(\mathbf{z}_{\mathrm{dmg}},\mathbf{z}_{\mathrm{ndmg}}) \gets f_\theta(\mathbf{X})$

where $\mathbf{z}_{\mathrm{dmg}} = [\mathbf{z}_{1}, \dots, \mathbf{z}_{N}]^\top \in \mathbb{R}^{N \times H}$

\vspace{0.3em}
\State \textbf{Step 2: Select baseline subset}
\State $N_b \gets \sum^N_{n=1}{b_n}$
\State $\mathbf{z}^b_{\mathrm{dmg}} \gets \{ \mathbf{z}_{n} \mid b_n = 1 \}$

where $\mathbf{z}^b_{\mathrm{dmg}} = [\mathbf{z}^b_{1}, \dots, \mathbf{z}^b_{N_b}]^\top \in \mathbb{R}^{N_b \times H}$

\vspace{0.3em}
\State \textbf{Step 3: Estimate baseline distribution}
\State $\boldsymbol{\mu} \gets \frac{1}{N_b} \sum^{N_b}_{n=1}\mathbf{z}^b_n$
\State $\boldsymbol{\Sigma}^{-1} \gets \left[\mathrm{Cov}(\mathbf{z}^b_{\mathrm{dmg}}) \right]^{-1}$

\vspace{0.3em}
\State \textbf{Step 4: Compute Mahalanobis damage scores}
\For{$n = 1$ to $N$}
    \State $m_n = (\mathbf{z}_n - \boldsymbol{\mu})^\top \boldsymbol{\Sigma}^{-1} (\mathbf{z}_n - \boldsymbol{\mu})$
\EndFor

\vspace{0.3em}
\State \textbf{Step 5: Determine detection threshold}
\State $\tau \gets \mathrm{Percentile}_p \left( \{ m_n \mid b_n = 1 \} \right)$

\vspace{0.3em}
\State \textbf{Step 6: Classify samples}
\For{$n = 1$ to $N$}
    \If{$m_n > \tau$}
        \State $\hat{y}_n \gets 1$ \Comment{damaged}
    \Else
        \State $\hat{y}_n \gets 0$ \Comment{undamaged}
    \EndIf
\EndFor

\State \Return $\{m_n\}_{n=1}^{N}, \tau, \{\hat{y}_n\}_{n=1}^{N}$

\end{algorithmic}
\end{algorithm}

\section{Real-world evaluation 1: openLAB bridge dataset}
This section presents the evaluation of the proposed structural damage identification framework using acceleration measurements from a full-scale girder bridge model, namely openLAB \cite{herbers2024openlab_paper}. The objective of this evaluation is to assess the effectiveness of the proposed method in a realistic civil infrastructure setting. Specifically, the following research questions are addressed:
\begin{enumerate}
    \item Can the proposed framework reliably detect and quantify structural damage under varying excitation and environmental conditions?
    \item Are all components and loss functions in the proposed deep learning architecture necessary for achieving robust performance?
    \item Can the proposed framework outperform traditional handcrafted feature-based methods for structural damage identification?
\end{enumerate}

\subsection{Dataset description}
The research bridge, openLAB, is a prestressed concrete bridge with a total length of approximately $3 \times 15$ m and a width of 4.5 m, located at the premises of Hentschke Bau GmbH in Bautzen, Germany \cite{herbers2024openlab_paper}. As illustrated in Figure \ref{fig:openlab}, the bridge consists of three spans. Spans 1 and 2 are continuous and feature a T-girder cross section, whereas span 3 is simply supported (structurally independent from spans 1 and 2) and has a hollow slab cross section. These structural configurations are representative of typical road bridges, making the dataset well suited for validating the proposed damage identification framework.

In this study, only spans 1 and 2 are considered, where artificial excitation and controlled damage were introduced. As shown in Figure \ref{fig:openlab}, six triaxial accelerometers (product name: Kistler K-Beam), labelled A1 to A6, are installed at the bottom of the mid-span cross sections of three T-girders on both spans. These sensors record vibration acceleration at a sampling frequency of 200 Hz.

\begin{figure*}[!htp] 
\centering
\includegraphics[width=12.1cm]{figures/openlab.png}
\caption{Overview of the openLAB bridge experiment. The top panel shows the full-scale bridge, while the middle panel illustrates the structural layout, including spans, shaker locations (S1–S2), accelerometer positions (A1–A6), the position of applied static load, and damage locations (D1–D3). The bottom panels present the cross sections of spans 1–2 (T-girder) and span 3 (hollow slab).}
\label{fig:openlab}
\end{figure*}

To introduce excitation variability, two electrodynamic shakers, labeled S1 and S2 in Figure \ref{fig:openlab}, are installed at the quarter-span locations of spans 1 and 2 on top of T-girder 1. The shakers generate non-stationary and broadband white-noise excitation in the frequency range of 1--80 Hz. By activating the shakers in different combinations, four excitation conditions are defined, as summarized in Table \ref{tab:exc_openlab}.

\begin{table}[!htp]
\centering
\caption{Excitation types of the openLAB dataset.}
\begin{tabular}{cl}
\toprule
\textbf{Label} & \textbf{Description} \\
\midrule
1 & Shaker 1 off, shaker 2 off (ambient excitation) \\
2 & Shaker 1 on, shaker 2 off \\
3 & Shaker 1 off, shaker 2 on \\
4 & Shaker 1 on, shaker 2 on \\
\bottomrule
\end{tabular}\\[10pt]
\label{tab:exc_openlab}
\end{table}

Regarding environmental conditions, the experiment was conducted from the afternoon of 5 May 2025 to the evening of 7 May 2025. According to historical weather records, all three days were sunny, with temperature ranging from 1 to 14\,\textdegree C and relative humidity ranging from 42\% to 88\%, providing environmental variability for this study. However, these environmental variables were not directly measured during the experiment.

To simulate structural damage, two types of interventions were applied. First, on 6 May, multi-step static loading was applied to span 2, girder 1 (see Figure \ref{fig:openlab} for the loading position) until different levels of cracks and permanent deformation were observed. However, the exact locations of the cracks were not recorded. Second, on 7 May, damage was introduced by drilling holes in the concrete and cutting prestressing strands. The corresponding locations are labelled D1 to D3 in Figure \ref{fig:openlab}. More information about the resulting damage states are summarized in Table \ref{tab:dmg_openlab}.

\begin{table}[!htp]
\centering
\caption{Damage states of the openLAB dataset.}
\begin{tabular}{cll}
\toprule
\textbf{Label} & \textbf{Time \& Date (CET)} & \textbf{Description} \\
\midrule
1 & \makecell[l]{from 13:30, 5 May \\ to 12:21, 6 May} & No damage \\
2 & \makecell[l]{from 12:21, 6 May \\ to 13:40, 6 May} & Crack width 0.1 mm \\
3 & \makecell[l]{from 13:40, 6 May \\ to 15:22, 6 May} & Crack width 0.25 mm \\
4 & \makecell[l]{from 15:22, 6 May \\ to 16:55, 6 May} & Crack width 0.3 mm \\
5 & \makecell[l]{from 16:55, 6 May \\ to 18:56, 6 May} & Crack width 0.4 mm \\
6 & \makecell[l]{from 18:56, 6 May \\ to 19:49, 6 May} & Crack width 0.5 mm \\
7 & \makecell[l]{from 19:49, 6 May \\ to 10:20, 7 May} & \makecell[l]{Permanent deformation \\ 5.7 mm} \\
8 & \makecell[l]{from 10:20, 7 May \\ to 14:50, 7 May} & Cut steel strands at D1 \\
9 & \makecell[l]{from 14:50, 7 May \\ to 18:03, 7 May} & \makecell[l]{Drill holes and cut \\ steel strands at D2} \\
10 & \makecell[l]{after 18:03, 7 May} & Cut steel strands at D3 \\
\bottomrule
\end{tabular}\\[10pt]
\label{tab:dmg_openlab}
\end{table}

During the experiment, vibration acceleration signals along with their UTC timestamps were recorded, resulting in a total duration of 340 minutes. The dataset covers various combinations of excitation conditions and damage states. Longitudinal acceleration is not considered in this study, as it is typically negligible for bridge structures. Consequently, a total of $6 \times 2 = 12$ sensor channels are used.

To facilitate model implementation, the continuous signals are segmented into 3400 overlapping windows, each with a length of 2048 data points (10.24 seconds) and an overlap of 1024 data points. The distribution of signal windows across different excitation conditions and damage states is summarized in Table \ref{tab:openlab_window}. Z-score normalization is applied to all windows, where, for each window, the mean and standard deviation computed across all sensor channels are normalized to 0 and 1, respectively. The PSD of each window is estimated using Welch’s method. Specifically, the \texttt{welch} function from the \texttt{scipy.signal} library is used with parameters \texttt{nperseg}=1024, \texttt{noverlap}=512, and \texttt{nfft}=2048, resulting in PSD vectors of length 1025 and a frequency resolution of 0.1953 Hz. The time series windows and their corresponding PSDs are split into training, validation, and test sets, containing 2040, 680, and 680 windows (0.6:0.2:0.2), respectively.

\begin{table}[!htp]
\centering
\caption{Number of signal windows from openLAB dataset.}
\begin{tabular}{cccccc}
\toprule
  & \textbf{Exc. 1} & \textbf{Exc. 2} & \textbf{Exc. 3} & \textbf{Exc. 4} & \textbf{Total} \\
\midrule
\textbf{Dmg. 1}  & 310 & 280 & 190 & 260 & 1040 \\
\textbf{Dmg. 2}  & 120 &  40 &  60 &  40 & 260 \\
\textbf{Dmg. 3}  &  30 &  40 &  50 &  40 & 160 \\
\textbf{Dmg. 4}  & 120 &  40 &  50 &  60 & 270 \\
\textbf{Dmg. 5}  & 120 &  40 &  50 &  50 & 260 \\
\textbf{Dmg. 6}  & 180 &   0 &   0 &   0 & 180 \\
\textbf{Dmg. 7}  & 180 &   0 &   0 &   0 & 180 \\
\textbf{Dmg. 8}  & 240 & 100 &  80 &  40 & 460 \\
\textbf{Dmg. 9}  & 240 &  50 &  90 &  50 & 430 \\
\textbf{Dmg. 10} & 120 &   0 &  40 &   0 & 160 \\
\textbf{Total}   & 1660 & 590 & 610 & 540 & 3400 \\
\bottomrule
\end{tabular}
\label{tab:openlab_window}
\end{table}

After signal segmentation and dataset splitting, all signal windows from 5 May in the training set are selected as the baseline subset for self-supervised learning. As shown in Table \ref{tab:dmg_openlab}, the bridge remains in a healthy state during both the afternoon of 5 May and the morning of 6 May. Nevertheless, only the data from the afternoon of 5 May are used as the baseline subset. This design enables us to evaluate whether the proposed framework can correctly classify the data from the morning of 6 May as healthy, despite differences in excitation and environmental conditions between the two periods. In addition, restricting the baseline subset to the training set prevents data leakage. The validation and test sets are not used to update the learnable parameters of the deep learning model, ensuring a fair and unbiased evaluation of its generalization performance.

Finally, it is worth noting that although parts of the openLAB dataset are publicly available \cite{herrmann_2026_openlab_data}, the data used in this study are not publicly accessible due to collaboration agreements with the commercial partner Kistler. More details can be found in the Data Availability Statement section at the end of this paper.

\subsection{Implementation details}
The proposed deep learning model is implemented in \texttt{PyTorch} 2.9.1 with \texttt{CUDA} 12.6 support. Model training is performed using the Adam optimizer with standard hyperparameters: a learning rate of 0.001, first- and second-moment decay coefficients $\beta_1 = 0.9$ and $\beta_2 = 0.999$, respectively, and a weight decay of 0.0001 to mitigate overfitting. All computations are executed on an NVIDIA GeForce RTX 4090 GPU for acceleration. The model is trained with a batch size of 256 for 500 epochs.

\begin{figure}[!htp] 
\centering
\includegraphics[width=8.5cm]{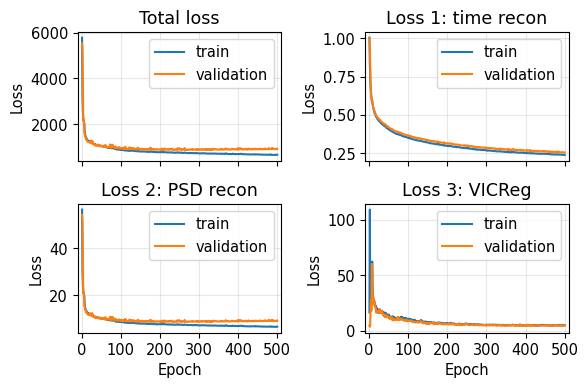}
\caption{Training and validation loss curves of the proposed framework on openLAB dataset. The total loss and individual loss components, including time-domain reconstruction (Loss 1), frequency-domain reconstruction (Loss 2), and VICReg self-supervised loss (Loss 3), are shown over 500 training epochs.}
\label{fig:openlab_loss}
\end{figure}

Both the encoder and Decoder 1 consist of four one-dimensional convolutional layers. The latent dimensions of $z_{\mathrm{dmg}}$ and $z_{\mathrm{ndmg}}$ are set to $H = 128$. The MLP used for PSD reconstruction comprises three layers with a hidden dimension of 512. The loss weighting coefficients are set as $\lambda_1 = 100$ for the time-domain reconstruction loss, $\lambda_2 = 100$ for the frequency-domain reconstruction loss, and $\lambda_3 = 1$ for the VICReg loss. These values are selected to yield good performance by balancing the contributions of the three loss terms, which differ in magnitude. Figure \ref{fig:openlab_loss} presents the training and validation loss curves from a representative run, with a total training time of 133.78 seconds. As observed, all loss components decrease smoothly during training, and the validation losses remain slightly higher than the corresponding training losses, indicating stable convergence and no evident overfitting.

The above settings are determined through preliminary hyperparameter tuning. It should be emphasized that the primary objective of this study is to demonstrate the feasibility and effectiveness of the proposed framework, rather than to present a fully optimized model. Although limited tuning has been conducted, extensive optimization is beyond the scope of this work. Interested readers are encouraged to further refine the model using the publicly available code and data provided in our GitHub repository \cite{jian2025ssrl}, which includes full implementation and configuration details.

\subsection{Evaluation results}
\begin{figure}[!htp] 
\centering
\begin{tabular}{c}
\includegraphics[width=8cm]{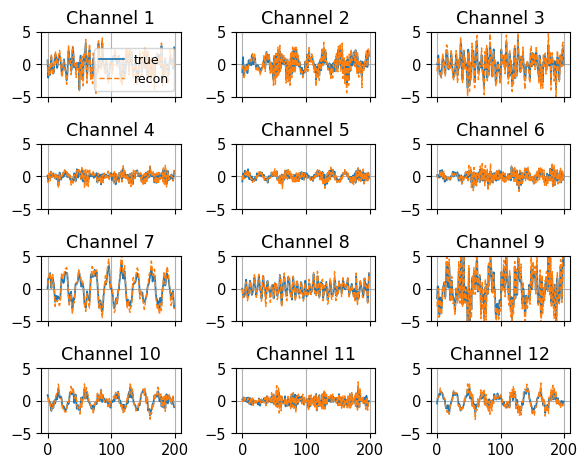} \\
(a) \\
\includegraphics[width=8cm]{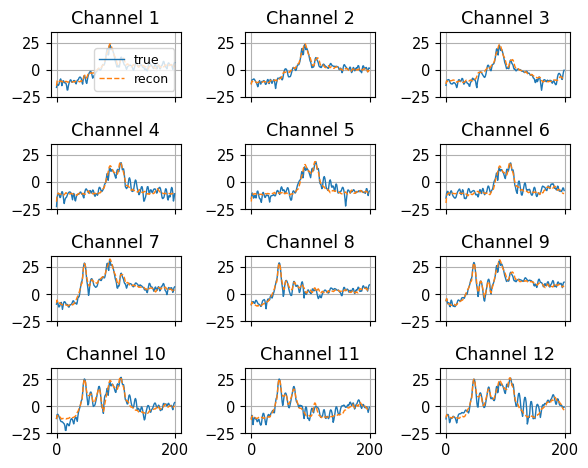} \\
(b) \\
\end{tabular}
\caption{Example reconstruction results for a signal window (damage label 1, excitation label 4). Only the first 200 data points are presented to allow clearer observation: (a) Time-domain acceleration signals and their reconstructions across 12 channels. (b) Corresponding PSD and their reconstructions.}
\label{fig:openlab_recon}
\end{figure}

After training, the model is evaluated on the validation and test set to assess its generalization capability, as these data are not used to update the learnable model parameters. The reconstructed acceleration time series and the corresponding PSD are first examined. Figure \ref{fig:openlab_recon} presents the reconstruction results for an example signal window. The close agreement between the original and reconstructed signals demonstrates the model’s ability to accurately capture both time-domain and frequency-domain characteristics.

\begin{figure}[!htp] 
\centering
\begin{tabular}{c}
\includegraphics[width=8cm]{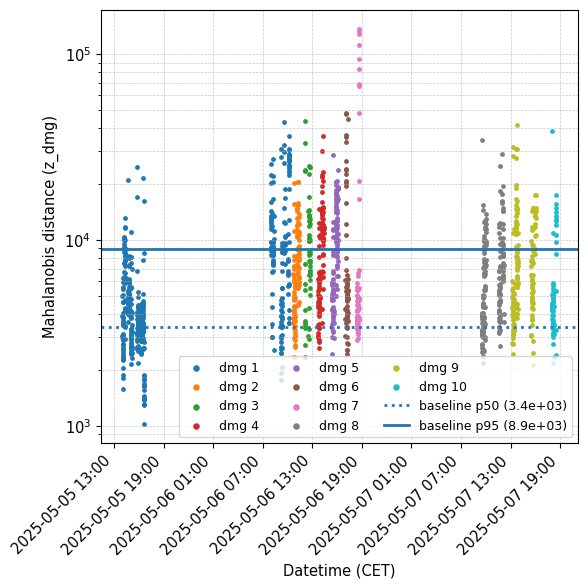} \\
(a) \\
\includegraphics[width=8cm]{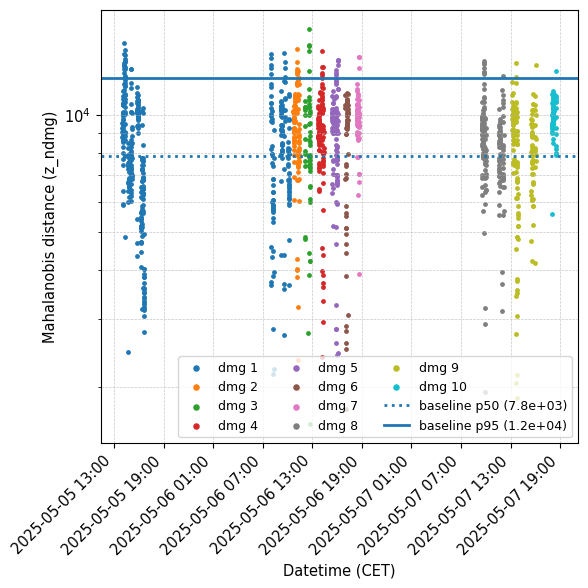} \\
(b) \\
\end{tabular}
\caption{Mahalanobis distances over time for all signal windows in the validation and test sets. Colours indicate different damage states. The dashed line denotes the baseline median, and the solid line denotes the 95th percentile threshold derived from the baseline subset: (a) Results from $z_\mathrm{dmg}$; (b) Results from $z_\mathrm{ndmg}$}
\label{fig:openlab_m_exc_all}
\end{figure}

The trained model is then applied to all signal windows in the validation and test sets. For each window, a Mahalanobis distance is computed based on the mean vector $\boldsymbol{\mu}$ and covariance matrix $\boldsymbol{\Sigma}$ estimated from the baseline subset used during training. Figure \ref{fig:openlab_m_exc_all} presents the scatter plot of the Mahalanobis distances for all signal windows, together with the median (50th percentile) and the 95th percentile (used as the damage detection threshold) of the Mahalanobis distances of the baseline subset. For comparison, both $z_\mathrm{dmg}$ and $z_\mathrm{ndmg}$ are used to compute the Mahalanobis distance.

Table \ref{tab:openlab_cm_exc_all} presents the confusion matrices for damage detection based on the results shown in Figure \ref{fig:openlab_m_exc_all}. In this study, the healthy state is defined as the negative class, while the damaged state is defined as the positive class. Accordingly, a true negative (TN) corresponds to a healthy signal window correctly classified as healthy, whereas a false positive (FP) denotes a healthy signal window incorrectly classified as damaged. Similarly, a false negative (FN) represents a damaged signal window incorrectly classified as healthy, and a true positive (TP) indicates a damaged signal window correctly classified as damaged.

Based on the confusion matrix, three evaluation metrics are used to assess the damage detection performance: the True Negative Rate (TNR), the True Positive Rate (TPR), and the balanced accuracy. These metrics are defined as follows:

\begin{equation}
\mathrm{TNR} = \frac{\mathrm{TN}}{\mathrm{TN} + \mathrm{FP}}
\end{equation}

\begin{equation}
\mathrm{TPR} = \frac{\mathrm{TP}}{\mathrm{TP} + \mathrm{FN}}
\end{equation}

\begin{equation}
\mathrm{Balanced\ Accuracy} = \frac{1}{2} \left( \mathrm{TPR} + \mathrm{TNR} \right)
\end{equation}

\begin{table}[!htp]
\centering
\caption{Confusion matrices and performance metrics for damage detection using Mahalanobis distance of all signal windows.}
\label{tab:openlab_cm_exc_all}
\resizebox{\linewidth}{!}{
\begin{tabular}{lccccccc}
\toprule
\textbf{} & \textbf{TN} & \textbf{FP} & \textbf{FN} & \textbf{TP} & \textbf{TNR} & \textbf{TPR}  & \textbf{\makecell[l]{Bal. \\ Acc.}} \\
\midrule
$\mathbf{z}_{\mathrm{dmg}}$  & 344 & 85  & 629 & 302 & 0.801 & \textbf{0.324} & \textbf{0.563} \\
$\mathbf{z}_{\mathrm{ndmg}}$ & 402 & 27  & 890 & 41  & \textbf{0.937} & 0.044 & 0.491 \\
\bottomrule
\multicolumn{8}{l}{\small *Metrics showing better performance are highlighted in bold.}
\end{tabular}
}
\end{table}

Results in Figure \ref{fig:openlab_m_exc_all} and Table \ref{tab:openlab_cm_exc_all} indicate the following conclusions:
\begin{enumerate}

\item As intended, the proposed framework effectively disentangles damage-related and non-damage-related information in the latent space of the autoencoder. The TPR values show that damage detection based on the damage-sensitive representation, $\mathbf{z}_{\mathrm{dmg}}$, significantly outperforms that based on $\mathbf{z}_{\mathrm{ndmg}}$. Although this improvement in TPR is accompanied by a reduction in TNR, this trade-off is acceptable in SHM practice, where detecting damage is of primary importance. Moreover, $\mathbf{z}_{\mathrm{dmg}}$ still achieves a higher balanced accuracy, indicating overall superior performance.

\item The overall detection accuracy is moderate. However, this outcome is both realistic and expected. In the openLAB experiment, the artificial damage was introduced only to girder 1 of span 2, which represents a relatively minor perturbation for a large-scale bridge structure. In addition, there are many signal windows under the weak ambient excitation. Since damage detection is performed at the level of individual signal windows, it is reasonable that such subtle damage is only detectable in a subset of windows where the non-stationary excitation sufficiently excites the affected structural components.

\item Damage quantification can be achieved by analysing the temporal evolution of the Mahalanobis distance. As shown in Figure \ref{fig:openlab_m_exc_all}(a), the Mahalanobis distances exhibit a gradual increasing trend over time, albeit with some variability. This trend is consistent with the progressive introduction of damage during the experiment.

\end{enumerate}

\begin{figure*}[!htp] 
\centering
\begin{tabular}{cccc}
(a) &\includegraphics[width=7cm]{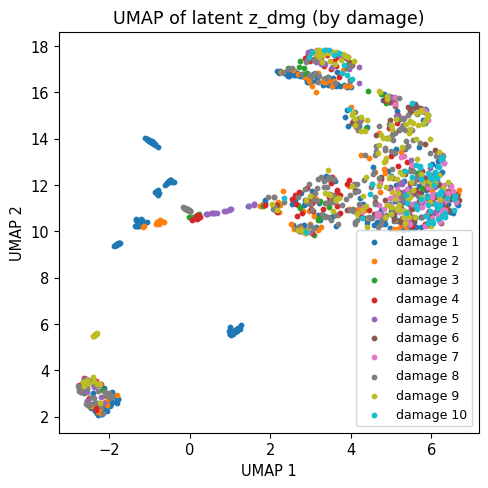} & (b) & \includegraphics[width=7cm]{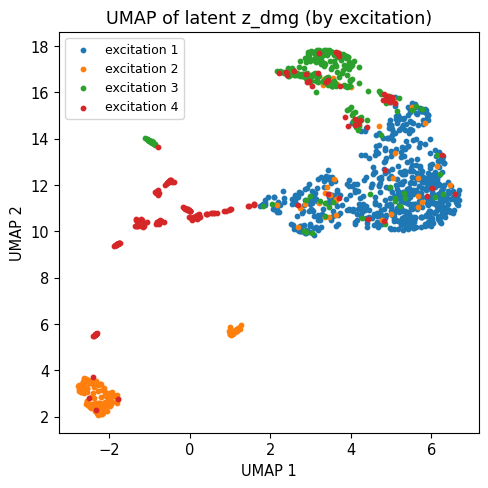} \\
(c) & \includegraphics[width=7cm]{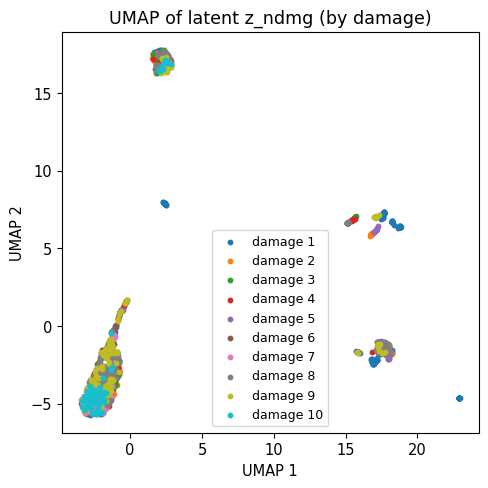} & (d) & \includegraphics[width=7cm]{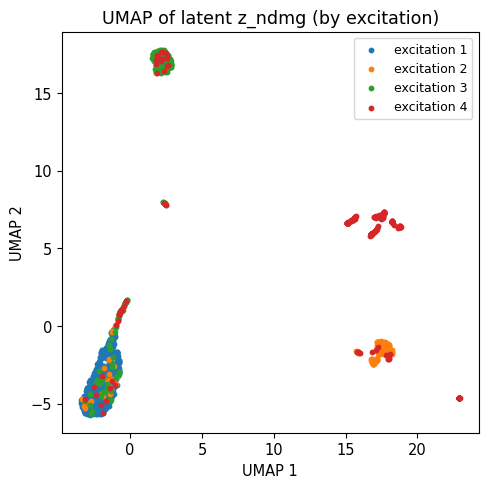} \\
\end{tabular}
\caption{UMAP visualization of the learned latent representations: (a) $\mathbf{z}_{\mathrm{dmg}}$ coloured by damage state; (b) $\mathbf{z}_{\mathrm{dmg}}$ coloured by excitation type; (c) $\mathbf{z}_{\mathrm{ndmg}}$ coloured by damage state; (d) $\mathbf{z}_{\mathrm{ndmg}}$ coloured by excitation type.}
\label{fig:openlab_umap}
\end{figure*}

To further demonstrate the disentanglement, the latent representations $\mathbf{z}_{\mathrm{dmg}}$ and $\mathbf{z}_{\mathrm{ndmg}}$ for all signal windows are visualized using UMAP \cite{mcinnes2018umap}. UMAP is a dimensionality reduction technique commonly used for visualization, similar to the classical t-SNE \cite{van2008visualizing}, but can offer clearer cluster separation. As shown in Figure \ref{fig:openlab_umap}:

\begin{enumerate}
\item The damage-related representation $\mathbf{z}_{\mathrm{dmg}}$ exhibits clearer separation across different damage states and weaker clustering with respect to excitation. In contrast, the nuisance representation $\mathbf{z}_{\mathrm{ndmg}}$ shows the opposite pattern. These results confirm that the proposed framework effectively disentangles damage-related and non-damage-related information.
\item Even so, the clusters corresponding to different damage states in Figure \ref{fig:openlab_umap} (a) are not completely separated. This is because, as discussed earlier, the artificial damage introduced in this experiment is relatively minor, making it insufficient to produce fully separable clusters across all damage states. This is also consistent with the results in Figure \ref{fig:openlab_m_exc_all}, where not all signal windows from damaged states are identified as damaged, and Mahalanobis distances of different damage states partly overlap. 
\end{enumerate}

Finally, Table \ref{tab:openlab_cm_diff_exc} presents the confusion matrices and corresponding damage detection metrics under different excitation conditions. It can be observed that the damage detection performance is the lowest under excitation label 1, which corresponds to weak ambient excitation. The performance improves significantly when external excitation from the shakers is introduced.

When only one shaker is active, the performance under excitation label 3 (shaker on span 2) is better than that under excitation label 2 (shaker on span 1). This can be explained by the fact that the structural damage is primarily located on span 2, making it more detectable when the excitation is applied closer to the damage location. The best performance is achieved under excitation label 4, where both shakers are active, providing the strongest excitation.

\begin{table}[!htp]
\centering
\caption{Confusion matrices and performance metrics for damage detection using Mahalanobis distance of signal windows under different excitation types.}
\label{tab:openlab_cm_diff_exc}
\resizebox{\linewidth}{!}{
\begin{tabular}{cccccccc}
\toprule
\textbf{Exc.} & \textbf{TN} & \textbf{FP} & \textbf{FN} & \textbf{TP} & \textbf{TNR} & \textbf{TPR}  & \textbf{\makecell[l]{Bal. \\ Acc.}} \\
\midrule
1  & 115 & 4  & 504 & 42  & 0.966 & 0.077 & 0.522 \\
2  & 89  & 31 & 61  & 46  & 0.742 & 0.430 & 0.586 \\
3  & 71  & 13 & 97  & 76  & 0.845 & 0.439 & 0.642 \\
4  & 68  & 38 & 29  & 76  & 0.642 & 0.724 & 0.683 \\
All  & 344 & 85 & 629 & 302 & 0.802 & 0.324 & 0.563 \\
\bottomrule
\end{tabular}
}
\end{table}

\begin{figure}[!htp] 
\centering
\includegraphics[width=8cm]{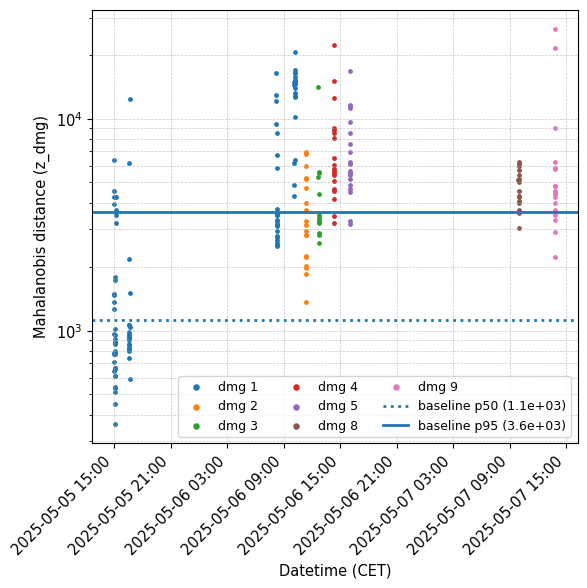} \\
\caption{Mahalanobis distances computed from $\mathbf{z}_\mathrm{dmg}$ for signal windows under excitation type 4. Colours indicate different damage states. The dashed line denotes the baseline median, and the solid line denotes the 95th percentile threshold derived from the baseline subset.}
\label{fig:openlab_m_exc4}
\end{figure}

Overall, these results indicate that stronger excitation leads to improved damage detectability. This observation is further supported by Figure \ref{fig:openlab_m_exc4}, which shows the Mahalanobis distances computed from $\mathbf{z}_{\mathrm{dmg}}$ for signal windows under excitation type 4 only. Under this strong excitation condition, a clear increasing trend in the Mahalanobis distance over time can be observed, reflecting the progressive introduction of damage during the experiment.

\subsection{Ablation study}
In this section, an ablation study is conducted to demonstrate the necessity of each component in the proposed model architecture. Five model variants are considered:

\begin{enumerate}
\item \textbf{Model F (Full model)}. The complete architecture shown in Figure \ref{fig:arch}, incorporating all three loss functions: time-domain reconstruction, self-supervised invariance (VICReg), and frequency-domain reconstruction.

\item \textbf{Model V1 (Autoencoder only)}. A baseline model trained using only the time-domain reconstruction loss (Loss 1), without any disentanglement or additional constraints.

\item \textbf{Model V2 (Autoencoder + self-supervision)}. A model trained using the time-domain reconstruction loss (Loss 1) and the self-supervised VICReg loss (Loss 3), aiming to enforce invariance to operational and environmental variations.

\item \textbf{Model V3 (Autoencoder + PSD reconstruction)}. A model trained using the time-domain reconstruction loss (Loss 1) and the frequency-domain reconstruction loss (Loss 2), incorporating spectral information but without explicit disentanglement.

\item \textbf{Model V4 (Handcrafted features)}. A non-deep-learning baseline that extracts statistical and spectral features directly from the input signals. Specifically, for each sensor channel, nine features are computed from the raw acceleration time series, including \textit{mean, standard deviation, skewness, kurtosis, maximum value, normalized peak frequency index, spectral entropy, normalized spectral centroid, and band power}. These features are then concatenated across all channels to form a feature vector for each signal window, leading to a total feature dimension of $9 \times 12 = 108$. This machine learning model serves as a conventional feature-engineering baseline to evaluate the benefit of the proposed deep learning framework.

\end{enumerate}

To evaluate the reliability and stability of the deep learning-based models, Model F and Models V1–V3 are each trained three times with different random seeds. The mean and standard deviation of their damage detection metrics are reported in Table \ref{tab:openlab_abalation}. Since Model V4 is based on handcrafted features and does not involve any training process, only a single set of metric values is reported.

\begin{table*}[!htp]
\centering
\caption{Damage detection performance of different model variants under various excitation conditions. Results for deep learning models are reported as mean $\pm$ standard deviation over three runs.}
\label{tab:openlab_abalation}

\begin{tabular}{ccccccc}
\toprule
\textbf{Model} & \textbf{Metric} & \textbf{Exc. 1} & \textbf{Exc. 2} & \textbf{Exc. 3} & \textbf{Exc. 4} & \textbf{All} \\
\midrule

\multirow{3}{*}{Model F} 
& TNR & 0.958$\pm$0.008 & 0.756$\pm$0.017 & 0.865$\pm$0.045 & 0.673$\pm$0.039 & 0.801$\pm$0.004 \\
& TPR & 0.079$\pm$0.006 & \textbf{0.417$\pm$0.057} & \textbf{0.347$\pm$0.087} & \textbf{0.673$\pm$0.114} & \textbf{0.319$\pm$0.007} \\
& Bal. Acc. & 0.519$\pm$0.003 & \textbf{0.587$\pm$0.022} & \textbf{0.606$\pm$0.049} & \textbf{0.673$\pm$0.039} & \textbf{0.560$\pm$0.005} \\

\midrule

\multirow{3}{*}{Model V1} 
& TNR & 0.966$\pm$0.000 & 0.756$\pm$0.013 & 0.948$\pm$0.007 & \textbf{0.981$\pm$0.009} & 0.952$\pm$0.009 \\
& TPR & 0.045$\pm$0.003 & 0.287$\pm$0.005 & 0.064$\pm$0.017 & 0.057$\pm$0.016 & 0.019$\pm$0.009 \\
& Bal. Acc. & 0.505$\pm$0.001 & 0.521$\pm$0.009 & 0.506$\pm$0.006 & 0.519$\pm$0.005 & 0.485$\pm$0.000 \\

\midrule

\multirow{3}{*}{Model V2} 
& TNR & 0.966$\pm$0.015 & \textbf{0.767$\pm$0.008} & 0.933$\pm$0.030 & 0.978$\pm$0.005 & \textbf{0.967$\pm$0.020} \\
& TPR & 0.062$\pm$0.023 & 0.277$\pm$0.005 & 0.064$\pm$0.026 & 0.010$\pm$0.010 & 0.011$\pm$0.005 \\
& Bal. Acc. & 0.514$\pm$0.005 & 0.522$\pm$0.002 & 0.498$\pm$0.010 & 0.494$\pm$0.005 & 0.489$\pm$0.008 \\

\midrule

\multirow{3}{*}{Model V3} 
& TNR & \textbf{0.975$\pm$0.008} & 0.761$\pm$0.017 & \textbf{0.952$\pm$0.012} & 0.811$\pm$0.081 & 0.912$\pm$0.027 \\
& TPR & 0.045$\pm$0.003 & 0.243$\pm$0.041 & 0.062$\pm$0.027 & 0.146$\pm$0.055 & 0.057$\pm$0.004 \\
& Bal. Acc. & 0.510$\pm$0.004 & 0.502$\pm$0.017 & 0.507$\pm$0.019 & 0.479$\pm$0.019 & 0.484$\pm$0.011 \\

\midrule

\multirow{3}{*}{Model V4} 
& TNR & 0.937 & 0.701 & 0.927 & 0.913 & 0.919 \\
& TPR & \textbf{0.112} & 0.261 & 0.086 & 0.101 & 0.108 \\
& Bal. Acc. & \textbf{0.525} & 0.481 & 0.506 & 0.507 & 0.513 \\

\bottomrule
\multicolumn{7}{l}{\small *Metrics showing better performance are highlighted in bold.}
\end{tabular}

\end{table*}

The results in Table \ref{tab:openlab_abalation} show that:
\begin{enumerate}
    \item Removing either the self-supervised invariance or the frequency-domain constraint leads to a significant drop in TPR, highlighting the importance of both components for robust damage-sensitive representation learning.
    \item Incorporating the self-supervised invariance and frequency-domain constraints leads to a reduction in TNR. However, this trade-off is expected and acceptable in the context of SHM, where missing damage (i.e., false negatives) is typically more critical than issuing false alarms. Importantly, Model F achieves the highest balanced accuracy among all variants, demonstrating that the overall detection performance is improved despite the decrease in TNR.
    \item The handcrafted feature-based variant (Model V4) is outperformed by Model F, but exhibits comparable performance to other deep learning variants (Models V1–V3). This suggests that deep learning does not inherently guarantee superior performance over conventional feature-based approaches. Instead, effective performance gains rely on carefully designed learning objectives and representations, particularly those that explicitly address nuisance variability and promote damage-sensitive feature extraction.
\end{enumerate}

\section{Real-world Evaluation 2: MCC5 gearbox dataset}
This section presents the evaluation of the proposed structural damage identification framework using a publicly available gearbox dataset, namely MCC5 \cite{chen2024multi}. The objective is to evaluate the generalization capability of the proposed method in a mechanical system under different operating conditions. The following research questions are investigated:
\begin{enumerate}
    \item Can the proposed framework reliably detect and quantify damage in mechanical systems under varying excitation conditions?
    \item Are all components and loss functions in the proposed deep learning architecture necessary for achieving robust performance?
    \item Can the proposed framework outperform traditional handcrafted feature-based methods for structural damage identification?
\end{enumerate}
\subsection{Dataset description}
\begin{figure*}[!htp] 
\centering
\begin{tabular}{cc}
\includegraphics[height=5cm]{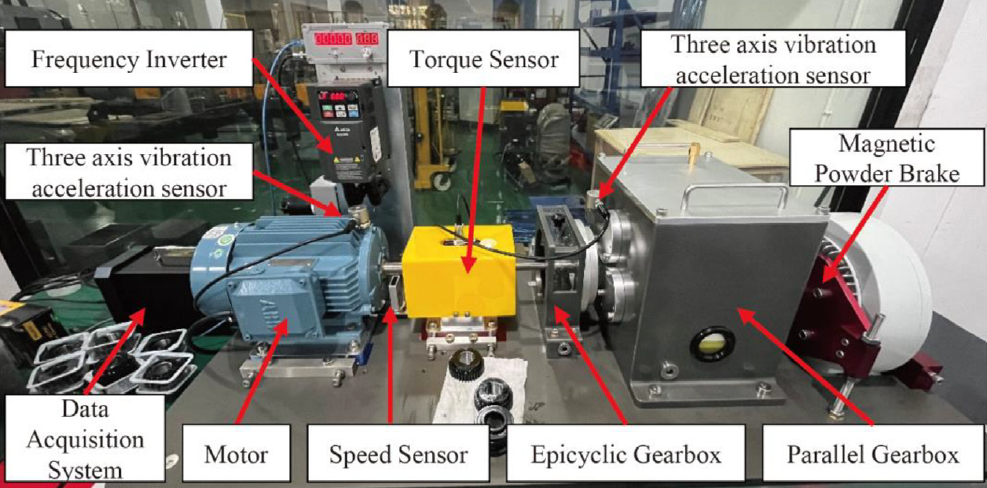} & \includegraphics[height=6cm]{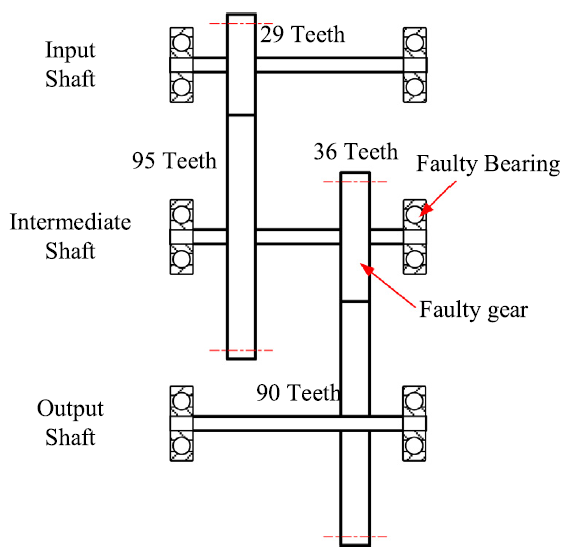} \\
(a) & (b) \\
\end{tabular}
\caption{Overview of the MCC5-THU gearbox experiment \cite{chen2024multi}: (a) Experimental setup of the MCC5-THU gearbox test rig, including the motor, torque sensor, parallel gearbox, magnetic powder brake, vibration sensors, and data acquisition system. (b) Schematic of the internal gearbox structure, showing the gear arrangement and the locations of artificially introduced gear and bearing faults.}
\label{fig:mcc5}
\end{figure*}

The MCC5-THU dataset is collected from gearboxes operating under variable working conditions, including time-varying rotational speed and load. The dataset comprises vibration acceleration signals, rotational speed measurements, and torque measurements, and covers a wide range of fault scenarios. These include multiple single-gear faults and combined bearing–gear faults. Figure \ref{fig:mcc5}(a) illustrates the experimental setup, which consists of a 2.2 kW three-phase asynchronous motor, a torque sensor, a two-stage parallel gearbox, a magnetic powder brake serving as a controllable load, and a data acquisition and control system. Figure \ref{fig:mcc5}(b) presents the internal configuration of the gearbox and the locations of the artificially introduced faults.

As shown in Figure \ref{fig:mcc5}(a), in this experiment four sensors are used, including a speed sensor that measures motor output shaft key phase signal, a torque sensor that measures gearbox input shaft torque, a three-axis accelerometer that measures the axial, horizontal, and vertical vibration acceleration of the motor drive end, and another three-axis accelerometer that measures the axial, horizontal, and vertical vibration acceleration of gearbox intermediate shaft bearing seat. Together there are 8 sensor channels, and the sampling frequency of all sensors is 12.8 kHz.

\begin{figure}[!htp] 
\centering
\includegraphics[width=0.9\linewidth]{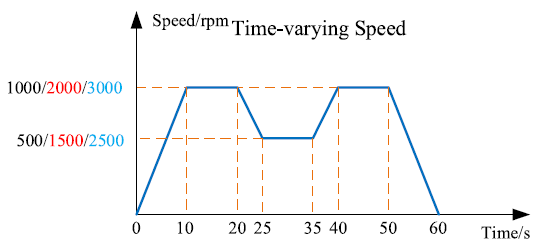}
\caption{Time-varying rotational speed profiles in the MCC5-THU dataset \cite{chen2024multi}. Each excitation sequence lasts 60 seconds and consists of piecewise speed levels with ramp transitions. Different colours indicate different experimental groups corresponding to distinct nominal rotational speeds.}
\label{fig:mcc5_exc}
\end{figure}

To introduce excitation variability, the time-varying shaft speed regime is considered in this study. The corresponding excitation profile is illustrated in Figure \ref{fig:mcc5_exc}. Each excitation sequence lasts 60 seconds and consists of multiple speed levels connected by ramp-up and ramp-down transitions. Different colours represent different experimental groups corresponding to distinct nominal rotational speeds. For example, the blue curve in Figure \ref{fig:mcc5_exc} shows one representative time-varying speed condition, where the rotational speed is increased from zero to a predefined level, maintained during specific time intervals (e.g., 10–20 s and 40–50 s), reduced to a lower level during 25–35 s, and finally ramped down to zero. 

Across different experiments, the predefined peak rotational speeds are set to 1000 rpm, 2000 rpm, and 3000 rpm. During all experiments considered in this study, the torque is kept constant of 10 Nm while only the rotational speed varies over time. Therefore, the excitation labels in this study are defined in Table \ref{tab:mcc5_exc}.

\begin{table}[!htp]
\centering
\caption{Excitation labels of the MCC5 dataset.}
\begin{tabular}{cl}
\toprule
\textbf{Label} & \textbf{Description} \\
\midrule
1 & Peak rotational speeds 1000 rpm \\
2 & Peak rotational speeds 2000 rpm \\
3 & Peak rotational speeds 3000 rpm \\
\bottomrule
\end{tabular}\\[10pt]
\label{tab:mcc5_exc}
\end{table}

No environmental variability is intentionally introduced in this experiment. The laboratory temperature is controlled within a narrow range of approximately $2\,^{\circ}\mathrm{C}$, ensuring stable environmental conditions throughout the data collection process.

The artificially introduced faults include both single gear faults and combined gear--bearing faults. All defects are created using laser etching with a machining precision of $0.01\,\mathrm{mm}$ to ensure repeatability and geometric accuracy. The detailed descriptions of the fault types and their corresponding parameters are summarized in Table~\ref{tab:mcc5_dmg}.

\begin{table}[!htp]
\centering
\caption{Damage states of the MCC5 dataset.}
\begin{tabular}{cll}
\toprule
\textbf{Label} & \textbf{Type} & \textbf{Description} \\
\midrule
1 & Healthy & No damage \\
2 & Gear pitting & Fault diameter $1.0\,\mathrm{mm}$ \\
3 & Gear wear & \makecell[l]{Wear over $1/2$ of \\ the tooth surface area}\\
4 & Missing tooth & One tooth removed \\
5 & \makecell[l]{Tooth breakage and \\ bearing inner fault} 
  & \makecell[l]{Inner raceway defect, \\ width $0.3\,\mathrm{mm}$} \\
6 & \makecell[l]{Tooth breakage and \\ bearing outer fault} 
  & \makecell[l]{Outer raceway defect, \\ width $0.3\,\mathrm{mm}$} \\
7 & Tooth breakage & \makecell[l]{Breakage over $1/2$ \\ of the tooth width} \\
8 & Tooth crack & \makecell[l]{Crack over $1/2$ \\ of the tooth height} \\
\bottomrule
\end{tabular}
\label{tab:mcc5_dmg}
\end{table}

During the experiment, data are collected under combinations of three excitation conditions and eight damage states, resulting in $3 \times 8 = 24$ operating conditions. For each condition, vibration signals are recorded for 60 seconds. Since this study focuses on output-only structural health monitoring, the speed and torque measurements are not used. Instead, only the six vibration channels, corresponding to the three-axis measurements of two accelerometers, are considered. 

The raw signals are downsampled from 12.8 kHz to 3.2 kHz. This downsampling is performed to balance computational efficiency and temporal resolution. If the original sampling frequency of 12.8 kHz were retained, a 1-second signal window would contain 12,800 data points, leading to excessive computational cost and memory consumption for the 1D CNN employed in this study. Reducing the sampling rate allows sufficiently long signal windows to be used for capturing meaningful temporal dynamics, while keeping the input dimensionality manageable for model training.

To facilitate model implementation, the continuous signals are segmented into 4320 overlapping windows, each with a length of 2048 data points (0.64 seconds) and an overlap of 1024 data points between adjacent windows. The distribution of signal windows across different excitation conditions and damage states is summarized in Table~\ref{tab:mcc5_window}. Z-score normalization is applied to all windows. For each window, the mean and standard deviation computed across all sensor channels are normalized to 0 and 1, respectively. The power spectral density (PSD) of each window is estimated using Welch’s method. Specifically, the \texttt{welch} function from the \texttt{scipy.signal} library is used with parameters \texttt{nperseg}=1024, \texttt{noverlap}=512, and \texttt{nfft}=2048, resulting in PSD vectors of length 1025 and a frequency resolution of 1.56~Hz. The time-series windows and their corresponding PSDs are randomly divided into training, validation, and test sets with a ratio of 0.6:0.2:0.2, containing 2592, 864, and 864 windows, respectively.

\begin{table}[!htp]
\centering
\caption{Number of signal windows from the MCC5-THU dataset.}
\begin{tabular}{ccccc}
\toprule
  & \textbf{Exc. 1} & \textbf{Exc. 2} & \textbf{Exc. 3} & \textbf{Total} \\
\midrule
\textbf{Dmg. 1}  & 180 & 180 & 180 & 540 \\
\textbf{Dmg. 2}  & 180 & 180 & 180 & 540 \\
\textbf{Dmg. 3}  & 180 & 180 & 180 & 540 \\
\textbf{Dmg. 4}  & 180 & 180 & 180 & 540 \\
\textbf{Dmg. 5}  & 180 & 180 & 180 & 540 \\
\textbf{Dmg. 6}  & 180 & 180 & 180 & 540 \\
\textbf{Dmg. 7}  & 180 & 180 & 180 & 540 \\
\textbf{Dmg. 8}  & 180 & 180 & 180 & 540 \\
\textbf{Total}   & 1440 & 1440 & 1440 & 4320 \\
\bottomrule
\end{tabular}
\label{tab:mcc5_window}
\end{table}

After signal segmentation and dataset splitting, all signal windows corresponding to the healthy state in the training set are selected as the baseline subset for self-supervised learning. Unlike the openLAB dataset, the MCC5 dataset does not contain time stamps, making it infeasible to define a baseline period based on chronological order. Therefore, the baseline subset must be defined based on a damage state that can reasonably be assumed to remain constant. In this study, the healthy state is chosen as the baseline reference. This choice is motivated by practical considerations: in real-world applications, damage typically accumulates progressively, and the healthy condition represents the nominal structural state prior to degradation. Importantly, damage labels are not used as supervision signals during model training. They are only employed to define a reference subset with approximately constant structural condition, which enables self-supervised invariance learning without introducing label information into the optimization process. Furthermore, restricting the baseline subset to the training set avoids data leakage and ensures an unbiased evaluation on the validation and test sets.

\subsection{Implementation details}
The implementation details of the proposed deep learning model on the MCC5 dataset are identical to those used for the openLAB dataset, as described in Section "Real-world validation 1: openLAB bridge dataset, Implementation details." For brevity, they are not repeated here.

\begin{figure}[!htp] 
\centering
\includegraphics[width=8.5cm]{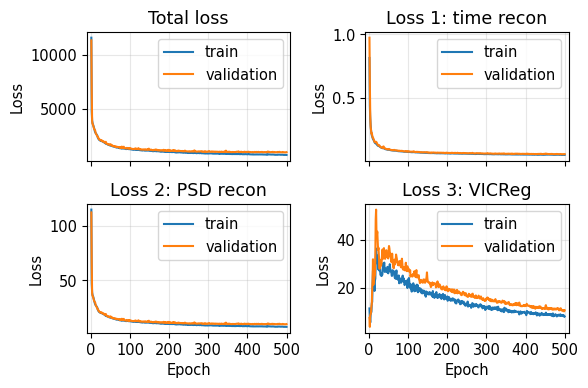}
\caption{Training and validation loss curves of the proposed framework on MCC5 dataset. The total loss and individual loss components, including time-domain reconstruction (Loss 1), frequency-domain reconstruction (Loss 2), and VICReg self-supervised loss (Loss 3), are shown over 500 training epochs.}
\label{fig:mcc5_loss}
\end{figure}

 Figure \ref{fig:mcc5_loss} presents the training and validation loss curves from a representative run, with a total training time of 161.02 seconds. As observed, all loss components decrease smoothly during training, and the validation losses remain slightly higher than the corresponding training losses, indicating stable convergence and no evident overfitting.

\subsection{Evaluation results}
\begin{figure}[!htp] 
\centering
\begin{tabular}{c}
\includegraphics[width=8cm]{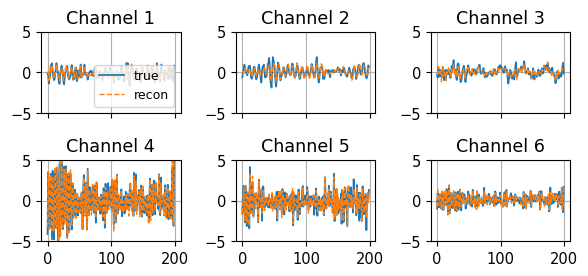} \\
(a) \\
\includegraphics[width=8cm]{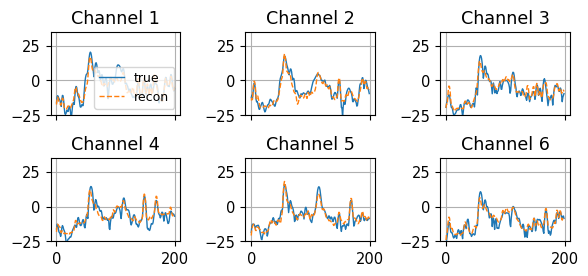} \\
(b) \\
\end{tabular}
\caption{Example reconstruction results for a signal window (damage label 3, excitation label 1). Only the first 200 data points are presented to allow clearer observation: (a) Time-domain acceleration signals and their reconstructions across 12 channels. (b) Corresponding PSD and their reconstructions.}
\label{fig:mcc5_recon}
\end{figure}

After training, the model is evaluated on the validation and test set to assess its generalization capability. The reconstructed acceleration time series and the corresponding PSD are first examined. Figure \ref{fig:mcc5_recon} presents the reconstruction results for an example signal window. The close agreement between the original and reconstructed signals demonstrates the model’s ability to accurately capture both time-domain and frequency-domain characteristics.

\begin{figure}[!htp] 
\centering
\begin{tabular}{c}
\includegraphics[width=8cm]{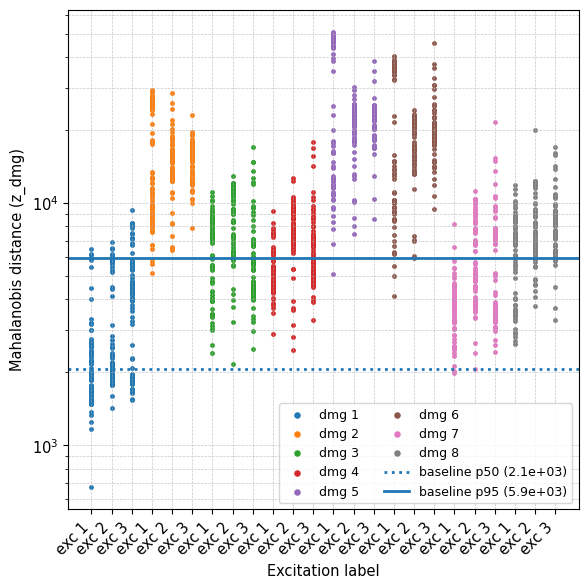} \\
(a) \\
\includegraphics[width=8cm]{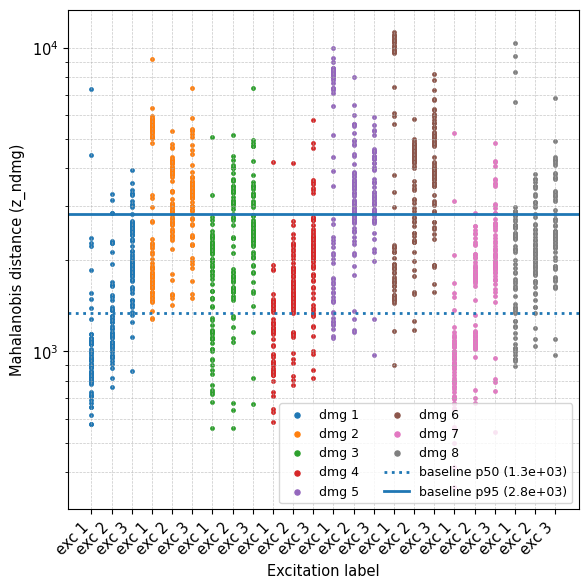} \\
(b) \\
\end{tabular}
\caption{Mahalanobis distances for all signal windows in the validation and test sets. Colours indicate different damage states. The dashed line denotes the baseline median, and the solid line denotes the 95th percentile threshold derived from the baseline subset: (a) Results from $z_\mathrm{dmg}$; (b) Results from $z_\mathrm{ndmg}$}
\label{fig:mcc5_m_exc_all}
\end{figure}

The trained model is then applied to all signal windows in the validation and test sets. For each window, a Mahalanobis distance is computed based on the mean vector $\boldsymbol{\mu}$ and covariance matrix $\boldsymbol{\Sigma}$ estimated from the baseline subset used during training. Figure \ref{fig:mcc5_m_exc_all} presents the scatter plot of the Mahalanobis distances for all signal windows, together with the median (50th percentile) and the 95th percentile (used as the damage detection threshold) of the Mahalanobis distances of the baseline subset. For comparison, both $z_\mathrm{dmg}$ and $z_\mathrm{ndmg}$ are used to compute the Mahalanobis distance.

Table~\ref{tab:mcc5_cm_exc_all} summarizes the confusion matrices for damage detection corresponding to the results in Figure~\ref{fig:mcc5_m_exc_all}. The notation and performance metrics follow those defined previously in Table~\ref{tab:mcc5_cm_exc_all}, and are not reiterated here to avoid redundancy.

\begin{table}[!htp]
\centering
\caption{Confusion matrices and performance metrics for damage detection using Mahalanobis distance of all signal windows.}
\label{tab:mcc5_cm_exc_all}
\resizebox{\linewidth}{!}{
\begin{tabular}{lccccccc}
\toprule
\textbf{} & \textbf{TN} & \textbf{FP} & \textbf{FN} & \textbf{TP} & \textbf{TNR} & \textbf{TPR}  & \textbf{\makecell[l]{Bal. \\ Acc.}} \\
\midrule
$\mathbf{z}_{\mathrm{dmg}}$  & 197 & 23  & 374 & 1134 & 0.896 & \textbf{0.752} & \textbf{0.824} \\
$\mathbf{z}_{\mathrm{ndmg}}$ & 204 & 16  & 963 & 545  & \textbf{0.927} & 0.361 & 0.644 \\
\bottomrule
\multicolumn{8}{l}{\small *Metrics showing better performance are highlighted in bold.}
\end{tabular}
}
\end{table}

Results in Figure \ref{fig:mcc5_m_exc_all} and Table \ref{tab:mcc5_cm_exc_all} indicate the following conclusions:
\begin{enumerate}

\item As intended, the proposed framework effectively disentangles damage-related and non-damage-related information in the latent space of the autoencoder. The TPR values show that damage detection based on the damage-sensitive representation, $\mathbf{z}{\mathrm{dmg}}$, significantly outperforms that based on $\mathbf{z}{\mathrm{ndmg}}$. Although this improvement in TPR is accompanied by a slight reduction in TNR, this trade-off is acceptable in SHM practice, where detecting damage is of primary importance. Moreover, $\mathbf{z}_{\mathrm{dmg}}$ still achieves a much higher balanced accuracy, indicating overall superior performance.

\item Compared with the damage detection results obtained from the openLAB dataset, the MCC5-THU dataset achieves substantially higher detection accuracy. This difference is expected because vibration signals from small mechanical systems, such as the gearbox used in the MCC5-THU experiment, are generally more sensitive to structural faults than those from large civil structures such as the openLAB bridge. In addition, differences in data acquisition conditions likely contribute to this performance gap. The MCC5-THU dataset appears to have been collected under controlled indoor conditions, resulting in reduced noise and fewer environmental variations (e.g., temperature fluctuations), whereas the openLAB bridge data was acquired in an outdoor environment where such factors introduce additional variability and measurement noise.

\item In addition to damage detection, partial damage quantification can be achieved by examining the Mahalanobis distances shown in Figure \ref{fig:mcc5_m_exc_all}(a). For example, the Mahalanobis distances corresponding to damage states 5 (tooth breakage with bearing inner race fault) and 6 (tooth breakage with bearing outer race fault) are noticeably larger than those of other damage states. This observation is consistent with the descriptions in Table \ref{tab:mcc5_dmg}, as damage states 5 and 6 involve combinations of multiple faults and are therefore expected to be more severe. However, this form of damage quantification is not fully reliable. For instance, damage state 2 (gear pitting) also exhibits relatively large Mahalanobis distances, even though its severity should be lower than that of damage states 3 (gear wear), 4 (missing tooth), 7 (tooth breakage), and 8 (tooth crack).

\item Although the Mahalanobis distance does not always follow the damage severity ranking, it can reveal differences in damage mechanisms. For example, gear pitting (damage state 2) produces noticeably larger Mahalanobis distances because localized pits generate repeated impulsive excitations during gear meshing, leading to strong broadband and high-frequency components that deviate significantly from the healthy baseline. In contrast, damage states 3, 4, 7, and 8 mainly introduce periodic stiffness modulation and sidebands around the gear-mesh frequency, resulting in more similar vibration characteristics and therefore comparable Mahalanobis distances. This behaviour is consistent with the proposed disentangled representation learning framework, where the latent space $\mathbf{z}{\mathrm{dmg}}$ is constrained by the PSD reconstruction loss.

\end{enumerate}

Finally, Table \ref{tab:mcc5_cm_diff_exc} presents the confusion matrices and corresponding damage detection metrics under different excitation conditions. Those results show that the proposed framework achieves consistently strong damage detection performance under different excitation conditions. The balanced accuracy ranges from 0.762 to 0.884 across the three excitation regimes, and remains high (0.823) when all excitation conditions are considered together. Among the three cases, excitation type 2 yields the best performance, suggesting that moderate excitation levels enhance the detectability of structural faults. In contrast, excitation type 3 leads to slightly reduced performance, likely due to increased operational variability that introduces additional nuisance factors. Nevertheless, the results demonstrate that the proposed framework maintains robust damage detection capability across varying excitation conditions.

\begin{table}[!htp]
\centering
\caption{Confusion matrices and performance metrics for damage detection using Mahalanobis distance of signal windows under different excitation types.}
\label{tab:mcc5_cm_diff_exc}
\resizebox{\linewidth}{!}{
\begin{tabular}{cccccccc}
\toprule
\textbf{Exc.} & \textbf{TN} & \textbf{FP} & \textbf{FN} & \textbf{TP} & \textbf{TNR} & \textbf{TPR} & \textbf{\makecell[l]{Bal. \\ Acc.}} \\
\midrule
1   & 66  & 6  & 116 & 372  & 0.917 & 0.762 & 0.839 \\
2   & 59  & 5  & 81  & 444  & 0.922 & 0.846 & 0.884 \\
3   & 70  & 14 & 152 & 343  & 0.833 & 0.693 & 0.763 \\
All & 197 & 23 & 374 & 1134 & 0.895 & 0.752 & 0.823 \\
\bottomrule
\end{tabular}
}
\end{table}

\subsection{Ablation study}
In this section, an ablation study is conducted again to verify the necessity of each component in the proposed architecture. The same five model variants defined in the openLAB study are considered here:

\begin{enumerate}
\item \textbf{Model F}: the full model with time-domain reconstruction, self-supervised invariance (VICReg), and frequency-domain reconstruction.

\item \textbf{Model V1}: autoencoder trained with time-domain reconstruction only.

\item \textbf{Model V2}: autoencoder with time-domain reconstruction and VICReg self-supervision.

\item \textbf{Model V3}: autoencoder with time-domain and frequency-domain reconstruction.

\item \textbf{Model V4}: a handcrafted feature benchmark using nine statistical and spectral features extracted from the vibration signals.
\end{enumerate}

The definitions of these variants are identical to those described in the openLAB section and are not repeated here for brevity.

To evaluate the reliability and stability of the deep learning-based models, Model F and Models V1–V3 are each trained three times with different random seeds. The mean and standard deviation of their damage detection metrics are reported in Table \ref{tab:mcc5_abalation}. Since Model V4 is based on handcrafted features and does not involve any training process, only a single set of metric values is reported.

\begin{table*}[!htp]
\centering
\caption{Damage detection performance of different model variants under various excitation conditions. Results for deep learning models are reported as mean $\pm$ standard deviation over three runs.}
\label{tab:mcc5_abalation}

\begin{tabular}{cccccc}
\toprule
\textbf{Model} & \textbf{Metric} & \textbf{Exc. 1} & \textbf{Exc. 2} & \textbf{Exc. 3} & \textbf{All} \\
\midrule

\multirow{3}{*}{Model F} 
& TNR & 0.917$\pm$0.000 & 0.917$\pm$0.024 & 0.877$\pm$0.038 & 0.911$\pm$0.016 \\
& TPR & \textbf{0.762$\pm$0.004} & \textbf{0.865$\pm$0.020} & \textbf{0.699$\pm$0.076} & \textbf{0.721$\pm$0.038} \\
& Bal. Acc. & \textbf{0.839$\pm$0.002} & \textbf{0.891$\pm$0.009} & \textbf{0.788$\pm$0.046} & \textbf{0.816$\pm$0.011} \\

\midrule

\multirow{3}{*}{Model V1} 
& TNR & 0.926$\pm$0.008 & 0.927$\pm$0.048 & 0.964$\pm$0.012 & 0.952$\pm$0.009 \\
& TPR & 0.076$\pm$0.004 & 0.099$\pm$0.053 & 0.042$\pm$0.009 & 0.062$\pm$0.003 \\
& Bal. Acc. & 0.501$\pm$0.005 & 0.513$\pm$0.010 & 0.503$\pm$0.006 & 0.507$\pm$0.005 \\

\midrule

\multirow{3}{*}{Model V2} 
& TNR & \textbf{0.954$\pm$0.029} & 0.964$\pm$0.009 & \textbf{0.972$\pm$0.014} & 0.947$\pm$0.014 \\
& TPR & 0.112$\pm$0.100 & 0.017$\pm$0.008 & 0.023$\pm$0.015 & 0.037$\pm$0.009 \\
& Bal. Acc. & 0.533$\pm$0.043 & 0.490$\pm$0.004 & 0.498$\pm$0.004 & 0.492$\pm$0.011 \\

\midrule

\multirow{3}{*}{Model V3} 
& TNR & 0.944$\pm$0.028 & \textbf{0.969$\pm$0.000} & 0.944$\pm$0.025 & \textbf{0.953$\pm$0.003} \\
& TPR & 0.083$\pm$0.048 & 0.040$\pm$0.002 & 0.052$\pm$0.022 & 0.062$\pm$0.007 \\
& Bal. Acc. & 0.514$\pm$0.016 & 0.504$\pm$0.001 & 0.498$\pm$0.018 & 0.508$\pm$0.005 \\

\midrule

\multirow{3}{*}{Model V4} 
& TNR & 0.950 & 0.950 & 0.950 & 0.950 \\
& TPR & 0.327 & 0.142 & 0.076 & 0.138 \\
& Bal. Acc. & 0.638 & 0.546 & 0.513 & 0.544 \\

\bottomrule
\multicolumn{6}{l}{\small *Metrics showing better performance are highlighted in bold.}
\end{tabular}

\end{table*}

The results in Table \ref{tab:mcc5_abalation} show that:
\begin{enumerate}
    \item The full framework (Model F) achieves the best overall performance across all excitation conditions. It consistently provides the highest TPR and balanced accuracy, indicating that combining time-domain reconstruction, VICReg self-supervision, and frequency-domain reconstruction enables the model to capture damage-related information more effectively than any partial variant.

    \item Models V1–V3 tend to achieve high TNR but very low TPR. This means that these simplified architectures can correctly recognize healthy states but fail to detect damaged states reliably. In other words, without the full set of constraints, the latent representation does not sufficiently emphasize damage-related features.

    \item The handcrafted feature baseline (Model V4) performs better than the simplified deep learning models in terms of balanced accuracy, but is still clearly inferior to Model F. This suggests that deep learning does not automatically outperform traditional feature engineering unless the model architecture is carefully designed.
\end{enumerate}

\section{Conclusions}
This study proposes a label-free self-supervised disentanglement learning framework for robust output-only vibration-based structural damage identification under operational variations. The framework is systematically evaluated on two structural vibration datasets: one from a full-scale bridge and the other from a high-fidelity gearbox. The key conclusions of this study are summarized as follows:

\begin{enumerate}

\item The proposed framework can successfully disentangle damage-related information from non-damage-related information. This is demonstrated by the fact that damage detection based on the latent representation $\mathbf{z}_\mathrm{dmg}$ significantly outperforms that based on $\mathbf{z}_\mathrm{ndmg}$. Using the disentangled damage-related representation, both damage detection and partial damage quantification can be achieved. The effectiveness of the framework is further supported by its successful application to two structurally different systems (a bridge and a gearbox), suggesting good generalizability.

\item The ablation study shows that successful disentanglement relies on all components of the proposed architecture. Removing any module leads to noticeable performance degradation. The proposed end-to-end framework also significantly outperforms the handcrafted feature benchmark, indicating that it can automatically extract useful information from vibration signals. Nevertheless, this does not imply that deep learning always outperforms handcrafted feature-based methods, as the handcrafted feature benchmark achieves performance comparable to several simplified deep learning variants.

\item Although the proposed framework can effectively disentangle damage-related information from nuisance factors, the damage detection performance is still influenced by structural scale and excitation characteristics. For example, the damage detection accuracy on the gearbox in the MCC5-THU dataset is higher than that on the bridge in the openLAB dataset, because vibration responses of small mechanical systems are generally more sensitive to structural faults than those of large civil structures. In addition, sufficiently strong and informative excitation improves damage detectability. This is supported by the openLAB results, where damage detection performance improves when the excitation changes from ambient excitation to shaker excitation, and when the shaker is located closer to the damage region.

\item Overall, the proposed framework provides a promising direction for robust and scalable vibration-based structural health monitoring, particularly in practical scenarios where damage labels are unavailable and operational conditions vary significantly.

\end{enumerate}

Finally, several directions for future work can be considered to further improve the proposed framework:

\begin{enumerate}

\item The evaluation results on the openLAB dataset show that damage detection accuracy is limited when the damage level is weak. To improve detection performance in such scenarios, more powerful representation learning models, such as transformer-based architectures or variational autoencoders, could be explored while maintaining the same self-supervised training strategy.

\item More advanced decision strategies could be adopted for damage detection. For example, instead of evaluating each signal window independently, one could compute the proportion of windows classified as damaged within a test sequence and apply a threshold to this ratio. Such approaches may improve robustness in weak-damage scenarios where only a subset of windows exhibit clear damage signatures.

\item The current damage quantification capability is still limited. Developing more reliable damage scoring methods or combining the learned representation with physics-informed indicators could further improve the reliability of damage quantification.

\end{enumerate}

\section{Data availability statement}
The code used to implement the proposed framework and reproduce the main results of this study is publicly available in our GitHub repository:
\href{https://github.com/JxdEngineer/SSRL}{https://github.com/JxdEngineer/SSRL}. The repository also includes detailed instructions and configuration files to facilitate reproduction of the reported experiments.
The openLAB dataset used in this work cannot be released due to a non-disclosure agreement with Kistler Instrumente AG. Nevertheless, a publicly available openLAB dataset collected with a different sensing system is available in this link \cite{herrmann_2026_openlab_data}, which can be used to reproduce similar experimental settings. The MCC5-THU dataset used in this study is publicly accessible and can be downloaded from the link included this paper \cite{chen2024multi}. 

\begin{dci}
The authors declared no potential conflicts of interest with respect to the research, authorship, and publication of this article.
\end{dci}

\begin{acks}
All model training and validation were conducted on the Euler cluster operated by the High Performance Computing group at ETH Z\"urich. The second author would like to acknowledge the Marie Skłodowska-Curie Actions (MSCA) program under the OMoRail project, Grant ID:~101104968.
\end{acks}

\bibliographystyle{SageV}
\bibliography{ref}

\end{document}